\pdfoutput=1

\documentclass[11pt]{article}

\usepackage[]{ACL2023}

\usepackage{enumitem}
\usepackage{comment}
\usepackage{times}
\usepackage{latexsym}
\usepackage{multirow}
\usepackage{graphicx}
\usepackage[T1]{fontenc}

\usepackage[utf8]{inputenc}

\usepackage{microtype}

\usepackage{inconsolata}

\NewDocumentCommand{\patricng}{ mO{} }{\textcolor{purple}{\textsuperscript{\textit{Patrick}}\textsf{\textbf{\small[#1]}}}}

\NewDocumentCommand{\zhiguow}{ mO{} }{\textcolor{blue}{\textsuperscript{\textit{Zhiguo}}\textsf{\textbf{\small[#1]}}}}
\NewDocumentCommand{\alex}{ mO{} }{\textcolor{red}{\textsuperscript{\textit{Alex}}\textsf{\textbf{\small[#1]}}}}
\NewDocumentCommand{\zshe}{ mO{} }{\textcolor{green}{\textsuperscript{\textit{Sheng}}\textsf{\textbf{\small[#1]}}}}
\NewDocumentCommand{\chaanj}{ mO{} }{\textcolor{magenta}{\textsuperscript{\textit{Anuj}}\textsf{\textbf{\small[#1]}}}}

\newcommand{\vspacereduce}{\vspace{-0.5em}}

\title{UNITE: A Unified Benchmark for Text-to-SQL Evaluation}

\author{%
Wuwei Lan, Zhiguo Wang, Anuj Chauhan, Henghui Zhu, Alexander Li, Jiang Guo, \\
\textbf{Sheng Zhang, Chung-Wei Hang, Joseph Lilien, Yiqun Hu, Lin Pan, Mingwen Dong,} \\
\textbf{Jun Wang, Jiarong Jiang, Stephen Ash, Vittorio Castelli, Patrick Ng and Bing Xiang}\\
  AWS AI Labs \\
  \texttt{\{lanwuwei, zhiguow, patricng\}@amazon.com}
 }

\begin{document}
\maketitle
\begin{abstract}
A practical text-to-SQL system should generalize well on 
a wide variety of natural language questions,
unseen database schemas, and novel SQL query structures. 
To comprehensively evaluate text-to-SQL systems, we introduce a \textbf{UNI}fied benchmark for \textbf{T}ext-to-SQL \textbf{E}valuation (UNITE). It is composed of publicly available text-to-SQL datasets, containing natural language questions from more than 12 domains, SQL queries from more than 3.9K patterns, and 29K databases. Compared to the widely used Spider benchmark \cite{yu-etal-2018-spider}, we introduce $\sim$120K additional examples and a threefold increase in SQL patterns, such as comparative and boolean questions.
We conduct a systematic study of six state-of-the-art (SOTA) text-to-SQL parsers on our new benchmark and show that: 1) Codex performs surprisingly well on out-of-domain datasets; 2) specially designed decoding methods (e.g. constrained beam search) can improve performance for both in-domain and out-of-domain settings; 3) explicitly modeling the relationship between questions and schemas further improves the Seq2Seq models.
More importantly, our benchmark presents key challenges towards compositional generalization and robustness issues --- which these SOTA models cannot  address well. \footnote{Our code and data processing scripts \url{https://github.com/awslabs/unified-text2sql-benchmark}}
\end{abstract}

\section{Introduction}

Text-to-SQL semantic parsing aims to transform natural language questions (NLQ) into SQL queries over underlying relational databases. It provides an interface for non-technical users to query their database. The current state-of-the-art text-to-SQL systems achieve more than 90\% accuracy on WikiSQL dataset \cite{zhongSeq2SQL2017} and 85\% accuracy on Spider dataset \cite{yu-etal-2018-spider}. However, these scores do not reflect real-world system performances. In real scenario, the text-to-SQL systems will encounter large database tables, out-of-domain entities, new compositional SQL structures, and diversified NLQ expressions. 
The existing single dataset is biased towards specific aspect and cannot evaluate text-to-SQL system comprehensively. For example, WikiSQL has only one table for each database, and the SQL statements have no JOIN clauses. Spider is known to have explicit column mention \cite{suhr-etal-2020-exploring} and simplified table names and column names \cite{lee-etal-2021-kaggledbqa}. The text-to-SQL datasets studied in \cite{finegan-dollak-etal-2018-improving} only have one database for each topic, which is inappropriate for cross-database evaluation. Spider-Syn \cite{gan-etal-2021-towards} is for robustness evaluation against synonym replacement, while Spider-DK \cite{gan-etal-2021-exploring} is designed for domain knowledge. Therefore, these datasets are proposed for a specific aspect of text-to-SQL task, the reported results in one or few selected datasets don't represent real-world performance.

In this work, we propose a \textbf{UNI}fied benchmark for \textbf{T}ext-to-SQL \textbf{E}valuation (UNITE), which consists of 18 publicly available text-to-SQL datasets covering different domains, including Wikipedia, healthcare, education, geography, transportation, software engineering, and finance. UNITE provides several benefits as a comprehensive text-to-SQL benchmark: 1) its dataset size is the largest to date, consisting of 97K training and 27K test examples; 2) it introduces a threefold increase of SQL patterns compared to the Spider benchmark, and contains examples from over 12 domains and 29K databases; and 3) its unified format aims to facilitate apples-to-apples text-to-SQL benchmarking on generalizability and robustness.

To better understand the challenges of UNITE benchmark, we evaluate six state-of-the-art text-to-SQL parsers. The best performance is below 50\%, showing the limited generalization ability of these text-to-SQL models in a realistic setting. Among these SOTA models, Codex \cite{codex} in-context learning \cite{codex_prompt} has the best out-of-domain performance, thus demonstrating the great potential of large-scale language models. 
We also observed that customized decoding, such as constrained beam search from PICARD \cite{scholak-etal-2021-picard},
and relation-aware self-attention can further improve the Seq2Seq models. 

In summary, our main contributions are: 
(1) We introduce a new benchmark with a wide variety of databases, SQL, and NLQ patterns to evaluate text-to-SQL comprehensively. 
(2) We perform in-depth statistics and analysis of our UNITE benchmark.
(3) We unify the format of 18 publicly available text-to-SQL datasets and remove the obstacles of data access and usage. We will release this unified collection and its evaluation framework to facilitate realistic text-to-SQL evaluation.
(4) We systematically study 6 SOTA text-to-SQL models (including Codex in-context learning) on our benchmark.

\section{UNITE Benchmark}

\subsection{Selection Criteria}
Our guideline of dataset selection is based on diversity, which means a dataset will be included in our benchmark if it can introduce new domains, new question types, new SQL patterns and new aspects for addressing specific issues.


\noindent\textbf{WikiSQL} \citet{zhongSeq2SQL2017} synthesized SQLs with pre-defined rules on top of Wikipedia tables, then generated crude NLQs from templates, which are rewritten and paraphrased by human annotators. 

\noindent\textbf{Spider} \citet{yu-etal-2018-spider} did not use any templates and directly instructed students from computer science major to write complex NLQ and SQL pairs over 200 databases. 

\noindent\textbf{SQUALL} To explore the necessity of fine-grained supervision for text-to-SQL semantic parsing, \citet{shi-etal-2020-potential} introduce detailed alignment between NLQ spans and SQL fragments (e.g., \textit{the highest} $\leftrightarrow$ \textit{ORDER BY ... LIMIT 1}) on top of WikiTableQuestions \cite{pasupat-liang-2015-compositional}.

\noindent\textbf{Spider-Syn} \citet{gan-etal-2021-towards} replace the schema-related words with the manually selected paraphrases, then eliminate the explicit mention between NLQ and schema to study model robustness against synonym substitution.

\noindent\textbf{Criteria2SQL} In medical domain, \citet{yu-etal-2020-dataset} introduce a text-to-SQL dataset that can translate eligibility criteria to executable SQL queries. All SQLs follow the same structure: SELECT \textit{patient$\_$id} FROM \textit{table$\_$id} WHERE condition.


\noindent\textbf{SparC} Instead of mapping a single complex question into a SQL query, \citet{yu-etal-2019-sparc} decomposes the long question in Spider into a sequence of simpler questions that are conditioned together to study the text-to-SQL task in context.

\noindent\textbf{CoSQL} \citet{yu-etal-2019-cosql} collect 3k dialogues through Wizard-or-Oz, where a DB user and a SQL expert interact with each other to finish the query process. The SQL expert needs to compose SQLs, retrieve answers, clarify questions and inform unanswerable questions.

\noindent\textbf{Spider-DK} \citet{gan-etal-2021-exploring} study the text-to-SQL robustness against rarely observed domain knowledge, where they define five types of domain knowledge and manually incorporate them into Spider development set.

\noindent\textbf{ParaphraseBench} \citet{paraphraseBench} study model robustness against different linguistic variations in NLQ, including naive, syntactic, morphological, lexical, semantic and missing information.

\noindent\textbf{XSP} \citet{suhr-etal-2020-exploring} repurpose eight text-to-SQL datasets cross database semantic parsing by filtering out redundant or non-trivial examples.
We select four of them that have a relatively large size
and use the original \textbf{Restaurants} dataset \cite{tang-mooney-2000-automated} instead, since it contains more novel SQL patterns.

\begin{table*}[ht]
\centering
\resizebox{\textwidth}{!}{
\begin{tabular}{lcccc|ccc|ccc|ccc}
\hline
\hline
\multirow{3}{*}{\textbf{Dataset}} & \multirow{3}{*}{\textbf{Domain}} &
\multicolumn{3}{c}{\textbf{Size}} &\multicolumn{3}{c}{\textbf{Schema}} &
\multicolumn{3}{c}{\textbf{NLQ}} &\multicolumn{3}{c}{\textbf{SQL}}  \\
 &  & \multirow{2}{*}{Train} & \multirow{2}{*}{Dev} & \multirow{2}{*}{Test} &  \multirow{2}{*}{\# DB} & \# Tables & \# Cols & Unique & \% of & \% of & Unique & Unique & Overlap \\
 & &  & & & & / DB & / Table & NLQs & COL & VAL & SQLs & Patterns &  w/ Spider\\
\hline
Spider & Misc. & 7000 & - & 1034$^*$ & 166 & 5.3 & 5.3 & 7990 & 49.1 & 95.8 & 4525 & 1009 & 100 \\
WikiSQL & Wikipedia & 56355 & 8421 & 15878 & 26531 & 1 & 7.3 & 80311 & 76 & 100 & 80175 & 556 & 4.3 \\
SQUALL & Wikipedia & 9069 & - & 2207$^*$ & 1617 & 1.9 & 9.5 & 11218 & 26 & 78.4 & 9814 & 687 & 16.2 \\
Spider-Syn & Misc. & 7000 & - & 1034 & 166 & 5.3 & 5.3 & 7995 & 35 & 95.4 & 4514 & 1009 & 98.4 \\
Criteria2SQL & Healthcare & 1400 & 300 & 303 & 371 & 1 & 11.7 & 1902 & 32.2 & 66.3 & 1515  & 366 & 5.2 \\
SparC & Misc. & 9025 & - & 1203$^*$ & 166 & 5.3 & 5.3 & 9161 & 50.4 & 96.2 & 8978 & 1385 & 67.6 \\
CoSQL & Misc. & 7343 & - & 1007$^*$ & 178 & 5.4 & 5.4 & 8187 & 54.6 & 94.2 & 8004 & 1316 & 46.5\\
Spider-DK & Misc. & - & - & 535 & 20 & 5.3 & 5.3 & 535 & 39.3 & 80.0 & 283 & 155 & 85.8 \\
ParaphraseBench & Medical & - & - & 342 & 1 & 1 & 8 & 57 & 75.9 & 92.2 & 56 & 42 & 47.6 \\
Restaurants & Restaurant & - & - & 309 & 1 & 3 & 4.3 & 125 & 0.5 & 81.6 & 23 & 16 & 0.0 \\
XSP-Advising & Education & - & - & 378 & 1 & 15 & 7.5 & 309 & 7.9 & 89.4 & 271 & 23 & 8.7  \\
XSP-ATIS & Transportation & - & - & 289 & 1 & 25 & 5.3 & 287 & 0.0 & 98.4 & 258 & 82 & 2.4 \\
XSP-GeoQuery & Geography & - & - & 532 & 1 & 7 & 4.3 & 532 & 23.3 & 98.1 & 363 & 75 & 22.7\\
XSP-IMDB & Entertainment & - & - & 107 & 1 & 17 & 4 & 107 & 4.8 & 100 & 103 & 41 & 7.3\\
KaggleDBQA & Kaggle & - & - & 272 & 8 & 2.1 & 11 & 272 & 29.0 & 80.2 & 249 & 120 & 50 \\
ACL-SQL & ACL & - & - & 491 & 1 & 13 & 2.2 & 465 & 37.0 & 97.5 & 445 & 56 & 21.4 \\
SEOSS-Queries & Software & - & - & 1162 & 1 & 13 & 6.2 & 1157 & 61.6 & 85.6 & 168 & 94 & 46.8 \\
FIBEN & Finance & - & - & 275 & 1 & 152 & 2.5 & 296 & 0.1 & 8.5 & 230 & 180 & 0.0 \\
\hline
\end{tabular}
}
\caption{\label{unified-benchmark-analysis}
Analysis of UNITE.
\# DB, \# Tables / DB and \# COLs / Table represent the number of databases, number of tables per database, and number of columns per table, respectively. For NLQ analysis, we measure the number of unique NLQs, the percentage of COLUMN mentions (\% of COL), and CELL VALUE mentions (\% of VAL) of the targeting SQLs. We also analyze SQL 
redundancy (unique SQLs and patterns) and novelty (extra new patterns introduced by measuring the overlap with Spider dataset).
$^*$~means original dev sets are used for test sets.
}
\vspacereduce
\end{table*}

\noindent\textbf{KaggleDBQA} \citet{lee-etal-2021-kaggledbqa} introduce a text-to-SQL dataset with realistic databases from the Kaggle website, these databases have abbreviated and obscure column names.

\noindent\textbf{ACL-SQL} On top of ACL anthology dataset \cite{singh-etal-2018-cl}, \citet{acl-sql} create ACL-SQL dataset that contain complex queries depending on up to five tables.

\noindent\textbf{SEOSS-Queries} \citet{seoss-queries} collect NLQs and SQLs from software engineering domain, then paraphrase with more variations. 

\noindent\textbf{FIBEN} \citet{fiben} introduce a text-to-SQL benchmark from financial domain, the NLQs are typical analytical queries generated by BI experts. Compared to other text-to-SQL datasets, the number of tables per database schema is at least an order of magnitude larger.

\subsection{Formatting}
Although these 18 text-to-SQL datasets are publicly available, it's non-trivial to convert them into the same format. For example, one of the SQUALL example has SQL label like this: `SELECT c1 FROM w ORDER BY c3\_number LIMIT 1', to make sure SQL target is aligned with natural language question, we recover SQL and database schema with the original column name and git rid of column variables (e.g. c1 and c3) for general purpose.
Finally, We convert all databases into SQLite format and NLQ/SQL pairs into JSONL format to simplify the training and evaluation process. For each example, there are three fields: database identifier, question and SQL query. For database schema, we keep both original names and clean names for tables and columns, we also keep column type, primary keys and foreign keys, all of them are saved in JSON object.

\subsection{Statistics and Analysis}
Table \ref{unified-benchmark-analysis} shows train/dev/test split of each dataset in our UNITE benchmark. To construct a cross-database setup, datasets with single database are used for test sets only. For the multi-database datasets, we follow the same split as the original work for fair comparison. Some datasets have no publicly available test sets, we use dev sets instead. The final benchmark has no database overlap between training and test set to satisfy the cross-database evaluation setup. 

We give detailed statistics from three aspects to demonstrate the diversity and comprehensiveness of our UNITE benchmark:

\noindent \textbf{Schema} Spider related datasets (e.g. Spider-Syn, SparC, Spider-DK) have similar database structure, for example, each database has $\sim$5 tables and each table has $\sim$5 columns on average. However, other databases are quite different, the table size ranges from 1 to 152, and the column size ranges from 2.5 to 11.7. These diversified schema will post new challenges for text-to-SQL systems.

\noindent \textbf{NLQ} Our benchmark contains various question types, such as \textit{When}, \textit{Who}, \textit{What}, \textit{How} and \textit{Which}, and also statement questions, comparative questions, and Boolean questions. Additionally, ParaphraseBench introduce different paraphrase types. The explicit column mentions and cell value mentions in NLQs imply the easiness  of producing the final SQL queries, since model can directly copy from NLQ to finish the column or cell value linking.
These two measurements are distinct across different datasets, and some of them are even lower than Spider, showing the potential challenges of our UNITE benchmark.

\noindent \textbf{SQL} Compositional generalization is critical to text-to-SQL semantic parsing. To measure this aspect, our UNITE benchmark contains various SQL patterns. Traditionally, Spider dataset is widely used for model training and evaluation, however, the distinct pattern overlap ratio in the last column of Table \ref{unified-benchmark-analysis} demonstrates the necessity of evaluating over a broader datasets. We also show SQL complexity measures in Appendix Table~\ref{unified-benchmark-analysis-appendix}, 
our benchmark consists of SQLs from different levels, ranging from single-table single-column to multi-table multi-column.

\begin{table*}[ht]
\centering
\footnotesize
\resizebox{\textwidth}{!}{
\begin{tabular}{lccccccccccc}
\hline
\hline
\multirow{3}{*}{\textbf{Dataset}} & \multicolumn{2}{c}{\textbf{Codex}} &
{\textbf{UL-20B}}&
\multicolumn{2}{c}{\textbf{RASAT}}&
{\textbf{SmBoP}}&
\multicolumn{2}{c}{\textbf{T5-3B}}&
\multicolumn{2}{c}{\textbf{PICARD}} & \multirow{3}{*}{\textbf{Prev.}}\\
 & zero & few & \multirow{2}{*}{Spider} & \multirow{2}{*}{Spider} & \multirow{2}{*}{UNITE} & \multirow{2}{*}{Spider} & \multirow{2}{*}{Spider} & \multirow{2}{*}{UNITE} & \multirow{2}{*}{Spider} & \multirow{2}{*}{UNITE} & \\
  & -shot & -shot & & & & & & & & &\\ 
\hline
Spider & 73.4 & 72.7 & 77.4 & 76.5 & 75.9 & 78 & 74.4 & 74.9 & \textbf{79.4} & 78.4 & 80.5 \\
WikiSQL &  44.9 & 46.3 & 38.4 & 20.2 & \textbf{84.8} & 32.7 & 24.1 & 84.1 & 40.3 & 84.2  & 92.7 \\
SQUALL & 50.0 & 51.6 & 28.2 & 15.6 & \textbf{71.4} & 25.2 & 13 & 71.1 & 19.7 & 64.6 & 42.2 \\
Spider-Syn & 61.7 & 43.8 & 72.7 & 62.2 & 70.1 & 65.4 & 63.1 & 70.0 & 70.9 & \textbf{74.0} & - \\
Criteria2SQL & 0 & 0 & 5.8 & 0 & \textbf{77.9} & 0 & 0 & 68.0 & 0 & 62.3 & 20.7 \\
SparC & 41.9 & 40 & 41.9 & 39.7 & 45.4 & \textbf{48.3} & 40.2 & 45.1 & 41.4 & 47.1 & 73.3\\
CoSQL  & 21.7 & 56.2 & 21.0 & 52.4 & \textbf{57.3} & 25.8  & 20.7 & 21.5 & 21.8 & 22.7 & 67.0 \\
\hline
\multicolumn{12}{c}{\textit{Out-of-domain}} \\
\hline
Spider-DK & \textbf{65.1} & 63.0 & 62.8 & 55.5 & 60.2 & 59.1 & 57.4 & 58.9 & 62.6 & 62.1 & - \\
ParaphraseBench & \textbf{88.3} & 88.0 & 78.4 & 75.1 & 81.6 & 80.7 & 82.5 & 82.8 & 81.6 & 83.0 & 74.6 \\
Restaurants & 39.4 & \textbf{51.1} & 41.3 & 36.5 & 34.9 & 18.3 & 23.8 & 30.2 & 46.0 & 47.6 & - \\
XSP-Advising & 7.1 & 6.8 & 7.4 & \textbf{11.7} & 10.4 & 1.9 & 6.8 & 9.1 & 6.8 & 9.7 & 6.9 \\
XSP-ATIS & 1.4 & \textbf{3.1} & 1.4 & 0.7 & 0.7 & 0 & 0.7 & 0.7 & 2.1 & 1.7 & 5.5\\
XSP-GeoQuery & 56.4 & 65.0 & 63.2 & 44.9 & 57.5 & 37.4 & 45.1 & 66.4 & 48.1 & \textbf{68.2} & 59.5 \\
XSP-IMDB & 37.4 & \textbf{41.1} & 24.3 & 24.3 & 26.2 & 18.7 & 21.5 & 6.5 & 28.0 & 7.5 & 37.1 \\
KaggleDBQA & 23.9 & \textbf{40.4} & 34.9 & 27.6 & 26.8 & 27.2 & 26.8 & 33.8 & 29.8 & 36.8 & 26.8 \\
ACL-SQL & \textbf{69.7} & 50.1 & 44.0 & 43.4 & 33.6 & 40.1 & 38.1 & 31.8 & 44.2 & 40.9 & - \\
SEOSS-Queries & \textbf{64.5} & 63.2 & 44.1 & 41.1 & 50.0 & 26.9 & 43.4 & 42.9 & 49.3 & 44.9 & -  \\
FIBEN & 0 & 0 & \textbf{28.4} & 0.3 & 4.4 & 5.1 & 0 & 2.0  & 11.5 & 26.7 & -\\
\hline
Average (overall) & 41.5 & 43.5 & 39.7 & 34.9 & \textbf{48.3} & 32.8 & 32.3 & 44.4 & 38.0 & 47.9 & - \\
Average (out-domain) & 41.2 & \textbf{42.9} & 38.9 & 32.8 & 35.1 & 28.7 & 31.5 & 33.2 & 37.3 & 39.0 & - \\
\hline
\end{tabular}
}
\caption{\label{unified-benchmark-experiments}
Comparison of different state-of-the-art models on UNITE. The last column "Prev." reports execution accuracy from previous works \cite{Qi2022RASATIR, xu-etal-2022-sead,shi-etal-2020-potential,yu-etal-2020-dataset,deng-etal-2021-structure}.
Columns \textit{Spider} and \textit{UNITE} 
represent models trained with Spider or UNITE, respectively. The \textit{overall} average refers to all test sets in our benchmark, while the \textit{out-domain} average considers datasets with test sets only.
}
\vspacereduce
\end{table*}

\subsection{Evaluation Metric}
Our UNITE benchmark contains SQLs from different standards, ground truth SQLs may have different realizations. Following previous work \cite{yu-etal-2018-spider, suhr-etal-2020-exploring}, we use execution accuracy to measure the correctness, which execute both predictions and ground truths over the underlying database and compare the execution results. We also filtered out some examples that have empty execution results and keep a relatively low ratio (e.g. $<$ 5\%), to accurately reflect denotation accuracy.


\section{Experiments}

We compare six state-of-the-art models in our experiments: \textbf{Codex} \citet{codex}, a 175B GPT model further fine-tuned on public code;
\textbf{UL-20B} \cite{ul2}, a 20B Seq2Seq model pretrained with mixture of denoisers; \textbf{T5-3B}, \citet{shaw-etal-2021-compositional} showed competitive performance of T5-3B for text-to-SQL task; \textbf{RASAT}, \citet{Qi2022RASATIR} further improved the T5-3B model by explicitly modeling the relationship between NLQs and schemas; \textbf{SmBoP} \cite{rubin-berant-2021-smbop}, a semi-autoregressive bottom-up parser that generates SQL tree reversely from leaf nodes to root node; \textbf{PICARD} \cite{scholak-etal-2021-picard}, a constrained beam search decoding method that generates SQL sequentially and discards illegal generations with incremental parsing. 

\noindent\textbf{Settings.} 
We report zero-shot and 3-shot 
performance for Codex, by following prompt design in \citet{codex_prompt}. For other fine-tuning based models, we report performance with the original public checkpoints trained on Spider dataset\footnote{UL-20B has no public checkpoint for text-to-SQL, we fine-tuned it with 7k training examples from Spider.}. We also re-train T5-3B and RASAT with our UNITE training set, which is a combination of the seven training sets in Table \ref{unified-benchmark-analysis}. We use codebase from previous works and keep hyperparameters unchanged. 


\subsection{Results and Analysis}
Our main results are shown in Table \ref{unified-benchmark-experiments}. Firstly, these SOTA models have performance level 70\%$\sim$80\% for Spider, but below 50\% on our UNITE benchmark on average. Thus the Spider benchmark is a poor proxy for evaluating real-world applications. Secondly, if we switch the training data from \textit{Spider} to \textit{UNITE}, we see consistent performance improvement from RASAT, T5-3B, and PICARD. This implies the necessity of more training instances for fine-tuning models. Thirdly,  Codex --- even without fine-tuning on any training data --- achieves the best out-of-domain performance. This is remarkable because very large language models pretrained on text and code can achieve similar or even better performance than models fine-tuned on rich human annotations. Other observations include:


\noindent \textbf{Codex is more robust to NLQ variations.} ParaphraseBench and SEOSS-Queries contain rich paraphrase variations, Codex performs better than fine-tuned models, we hypothesize that the improvements are attributed to (1) model size, i.e. Codex has 175B parameters, but others have at most 20B; and (2) learning paradigm, i.e. in-context learning ingests data in a more natural way and does not update model parameters.

\noindent \textbf{Codex zero-shot sometimes outperforms the few-shot setting.} Codex\footnote{Note that Codex may have seen some of the evaluation sets via webscraping or GitHub during pretraining.}  zero-shot has a lower average score, but it outperforms few-shot on 8 out of 18 datasets. We hypothesize that prompt engineering or picking the correct exemplars is important to in-context model performance on out-of-domain data.



\noindent \textbf{Customized decoding is important for text-to-SQL generalization.} The PICARD model improves T5-3B model on 15 out of 22 datasets and increases both in-domain and out-domain performance. The SmBoP model, even with a smaller Grappa encoder than T5-3B, also shows advantage on some datasets, such as Spider, WikiSQL, SQUALL and SparC.

\noindent \textbf{Relation-aware self-attention further improves the Seq2Seq model.} Comparing T5-3B and RASAT, we can see the benefit of explicitly modeling the relational structures for schema linking and schema encoding, which is realized through self-attention mechanism in the pre-trained T5 model.


\section{Conclusion}
We proposed a UNITE benchmark to evaluate text-to-SQL in a realistic setting, which consists of 18 datasets that cover a broad range of databases, SQLs and NLQs, we compared six SOTA models systematically and analyzed their performance with regarding to model architecture and decoding strategies. We will make our benchmark publicly available to benefit the whole community.

\section*{Limitations}
UNITE is a not a multi-lingual text-to-SQL benchmark, we only consider English and may lose opportunities to other languages. In our experiments, we did not retrain UL-20B with UNITE training set because of limited compute resources, therefore we miss the best performance of fine-tuning based models. Since some models (e.g. Codex, UL-20B and SmBoP) are not designed for conversational text-to-SQL, to fairly compare all the SOTA models, we didn't consider conversation history in our experiments, therefore we see lower performance on SparC and CoSQL compared to previous works. Considering model size and limited compute, we don't run experiments with multiple seeds.


\section*{Ethics Statement}
Our benchmark experiments take extremely high amount of compute resources, which may not be environment-friendly. For example, UL-20B fine-tuning experiment needs 64 NVIDIA A100-SXM4-40GB GPUs for a total of 10 days training time. Our UNITE training set is 13 times larger than Spider, which can increase the training time and cause more CO2 emission, for example, T5-3B needs two more days to get converged in a single A100-SXM4-40GB GPU card.
\bibliography{anthology,custom}
\bibliographystyle{acl_natbib}

\clearpage
\appendix

\section{Appendix}\label{sec:appendix}

\begin{table}[ht]
\centering
\footnotesize
\resizebox{0.48\textwidth}{!}{
\begin{tabular}{lcccc}
\hline
\hline
\multirow{3}{*}{\textbf{Dataset}} & \multicolumn{4}{c}{\textbf{SQL}}  \\
 & \# JOINs & Nesting & Unique & \# SQLs \\
 & / Query & Depth & Patterns &  / Pattern\\
\hline
Spider  & 0.5 &1.1 & 1009 & 8 \\
WikiSQL & 0 & 1 & 556 & 145.1 \\
SQUALL  & 0.1 & 1.2 & 687 & 16.4 \\
Spider-Syn & 0.5 & 1.1 & 1009 & 8 \\
Criteria2SQL & 0 & 1  & 366 & 5.5 \\
SparC & 0.5 & 1.1 & 1385 & 7.4 \\
CoSQL & 0.4 & 1.1 & 1316 & 6.3\\
Spider-DK  & 0.6 & 1.1 & 155 & 3.5 \\
ParaphraseBench & 0 & 1 & 42 & 1.4 \\
Restaurants & 0 & 1.3 & 16 & 23.6 \\
XSP-Advising & 0.4 & 1.1 & 23 & 13.4  \\
XSP-ATIS & 0 & 1.1 & 82 & 3.5 \\
XSP-GeoQuery  & 0.01 & 1.6 & 75 & 7.1\\
XSP-IMDB  & 0 & 1 & 41 & 2.6\\
KaggleDBQA & 0.2 & 1.1 & 120 & 2.3 \\
ACL-SQL & 1.1 & 1 & 56 & 8.8 \\
SEOSS-Queries  & 0.3 & 1.1 & 94 & 12.4 \\
FIBEN  & 4.0 & 1.5 & 180 & 1.6 \\
\hline
\end{tabular}
}
\caption{\label{unified-benchmark-analysis-appendix}
SQL complexity (\# JOINs per query and nesting depth) and redundancy (\# SQLs per pattern) analysis on UNITE benchmark.
}

\end{table}
As shown in Table \ref{unified-benchmark-analysis-appendix}, our UNITE benchmark contains datasets that have more complex SQLs than Spider data, for example, FIBEN has four JOINs per SQL query, XSP-GeoQuery has 1.6 nesting depth\footnote{The nesting depth is defined as Count of SELECT - (Count of INTERSECT + Count of EXCEPT).}. We also have simple SQLs with only one table involved, such as WikiSQL, Criteria2SQL, ParaphraseBench, etc. 

\end{document}